\documentclass{article}

\usepackage[preprint,nonatbib]{neurips_2025}

\usepackage[utf8]{inputenc} 
\usepackage[T1]{fontenc}    
\usepackage{hyperref}       
\usepackage{url}            
\usepackage{booktabs}       
\usepackage{amsfonts}       
\usepackage{nicefrac}       
\usepackage{microtype}      
\usepackage{xcolor}         

\usepackage{amsmath}
\usepackage{soul}
\usepackage{graphicx}
\usepackage{algorithm}
\usepackage{algpseudocode}
\usepackage{caption}
\usepackage{multirow}
\usepackage{subcaption}
\usepackage[utf8]{inputenc}
\usepackage{amsmath}
\usepackage{booktabs}
\usepackage{amsfonts}

\title{TinyDrive: Multiscale Visual Question Answering with Selective Token Routing for Autonomous Driving}

\author{%
  Hossein Hassani\\
  Western University\\
  \texttt{hhassa52@uwo.ca} \\
  \And
  Soodeh Nikan \\
  Western University\\
  \texttt{snikan@uwo.ca} \\
  \AND
  Abdallah Shami \\
  Western University\\
  \texttt{abdallah.shami@uwo.ca} \\
}

\begin{document}

\maketitle

\begin{abstract}
Vision Language Models (VLMs) employed for visual question-answering (VQA) in autonomous driving often require substantial computational resources that pose a challenge for their deployment in resource-constrained vehicles. To address this challenge, we introduce TinyDrive, a lightweight yet effective VLM for multi-view VQA in driving scenarios. Our model comprises two key components including a multiscale vision encoder and a dual-level prioritization mechanism for tokens and sequences. The multiscale encoder facilitates the processing of multi-view images at diverse resolutions through scale injection and cross-scale gating to generate enhanced visual representations. At the token level, we design a token routing mechanism that dynamically selects and process the most informative tokens based on learned importance scores. At the sequence level, we propose integrating normalized loss, uncertainty estimates, and a diversity metric to formulate sequence scores that rank and preserve samples within a sequence priority buffer. Samples with higher scores are more frequently selected for training. TinyDrive is first evaluated on our custom-curated VQA dataset, and it is subsequently tested on the public DriveLM benchmark, where it achieves state-of-the-art language understanding performance. Notably, it achieves relative improvements of $11.1\%$ and $35.4\%$ in BLEU-4 and METEOR scores, respectively, despite having a significantly smaller parameter count.
\end{abstract}

\section{Introduction}\label{secI}
The current Autonomous Driving Systems (ADS) are capable of accurately perceiving and navigating their environments \cite{hassani2024traffic}. The future ADS, however, are rapidly moving toward intelligent systems that not only understand their environments but also reason about their decisions in a human-like manner. As ADS evolve toward higher levels of autonomy, there is a pressing need for models that go beyond the basic interpretation of raw sensory measurements in order to provide semantically rich, explainable insights for real-time decision-making \cite{guan2024world}. In this respect, Vision Language Models (VLMs) have shown a promising perspective by augmenting visual perception into natural language understanding, which ultimately forms a structured interface between the ADS and the downstream decision-making module \cite{zhou2024vision}. The overarching goal is empowering the vehicle to meaningfully interpret and communicate the context of its immediate surroundings. This is achievable by means of VLMs because they leverage a vision encoder that extracts visual features and a Large Language Model (LLM) that reasons about them to generate the response through an interactive Visual Question-Answering (VQA).

Even though VLMs show strong performance under open-domain VQA tasks, they suffer from high computational overhead. This is because the current models used for VQA in autonomous driving rely on large pretrained vision encoders and LLMs. For example, models such as DriveMLM \cite{wang2023drivemlm}, Drive-GPT4 \cite{xu2024drivegpt4}, LLM-Driver \cite{chen2024driving}, and DriveLM-Agent \cite{sima2024drivelm} have billions of parameters mainly due to relying on Vision Transformer (ViT)-based vision encoders as well as costly fusion of vision and text embeddings. More recently, $\text{EM-VLM4AD}_{\text{Q-Large}}$ (with $769$M trainable parameters) and $\text{EM-VLM4AD}_{\text{Base}}$ (with $235$M parameters) \cite{gopalkrishnan2024multi} models with less than a billion trainable parameters proposed a Gated Pooling Attention (GAP) to unify multi-view vision embeddings and connect them with the text embeddings. Building on top of the EM-VLM4AD, MiniDrive \cite{zhang2024minidrive} is constructed based upon Convolution Neural Networks (CNNs) with large kernels (UniRepLKNet) \cite{ding2024unireplknet} as the backbone vision encoder, where through a mixture-of-experts, they generate $2$D visual features. They also used cross attention to fuse the extracted visual features with text embeddings as input to the Text-to-Text Transfer Transformer (T5) \cite{raffel2020exploring} language model. The smallest version of MiniDrive with $83$M parameters has achieved competitive performance with models having billions of parameters. The attained results show that CNN-based vision encoders can not only reduce computational overhead but also improve overall performance. Inspired by this, our work is devoted to designing a compact, CNN-based VLM that achieves a better trade-off between computational complexity and performance for VQA in autonomous driving.

Our model, termed \textit{TinyDrive}, incorporates a multiscale CNN-based vision encoder, a token routing mechanism, and a sequence priority buffer to construct a compact VLM. We develop a mutiscale CNN to capture fine-grained as well as contextual features across various resolutions (high, medium, and low) in order to ensure a comprehensive visual understanding. This vision backbone is then followed by a Local-Global Block (LGB), which further processes the features before passing them to a classification head. This head is used to generate control commands for a Yahboom Rosmaster X3 self-driving car used for constructing a custom VQA dataset. Within the token routing mechanism, we select and process only informative text tokens based on a learned score. This score comprises three factors: the embedding magnitude, their positions within the underlying sequence, and a padding mask. We then retain the top $K$ text tokens and less informative embeddings are masked and pruned from further processing. These selected tokens and vision embeddings are then further assessed by a priority score based on a combination of loss gradients, uncertainty estimates, and a diversity metric. Using the priority score, we retain and rank sequences within a sequence priority buffer. The likelihood of a sequence to be selected for fine-tuning the language model is directly proportional to its overall score. This buffer aims to prioritize highly informative sequences during training. 

\begin{figure}[htbp]
    \centering
    \subfloat[]{\includegraphics[width=0.3\textwidth]{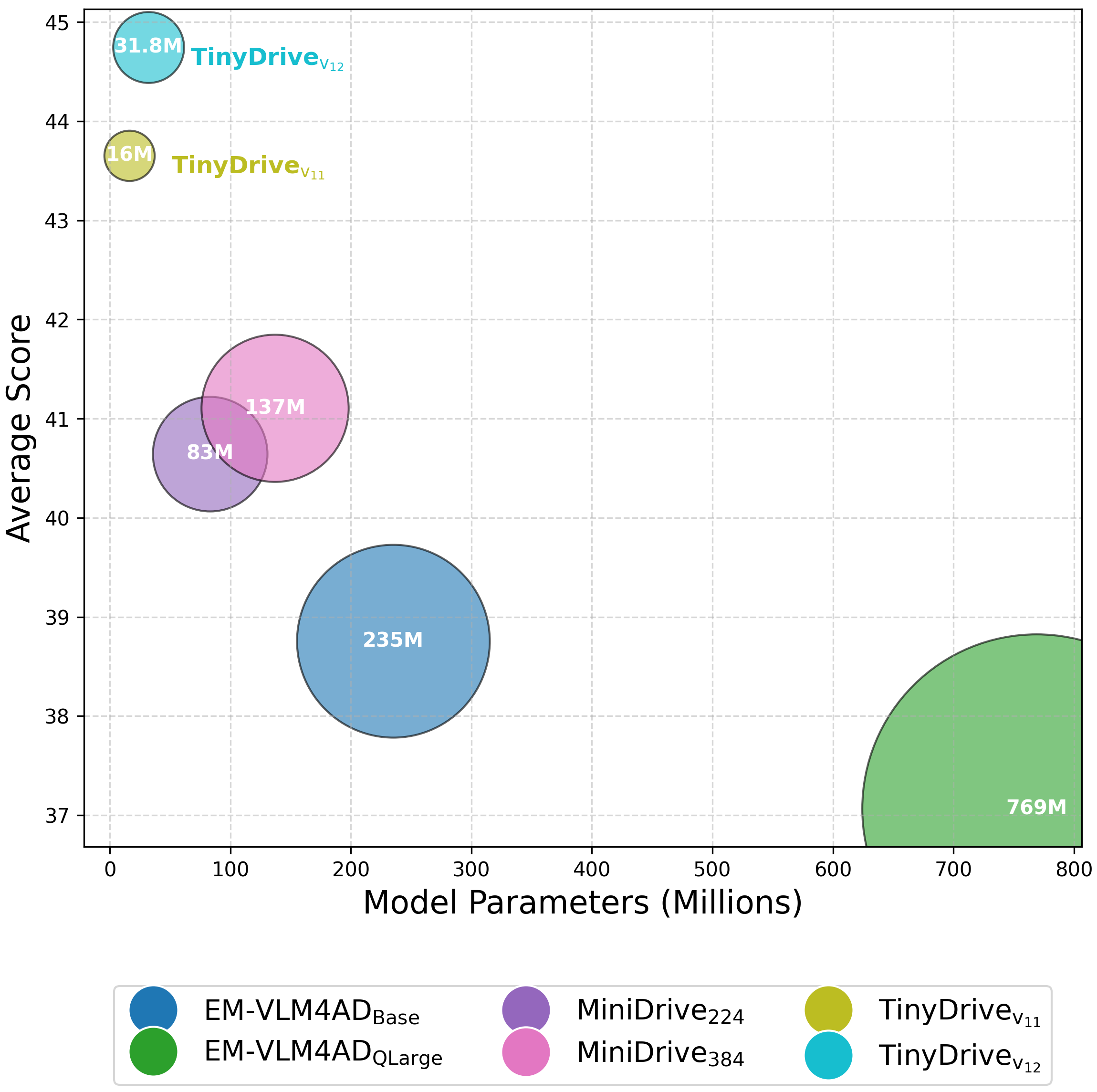}}
    \quad
    \subfloat[]{\includegraphics[width=0.3\textwidth]{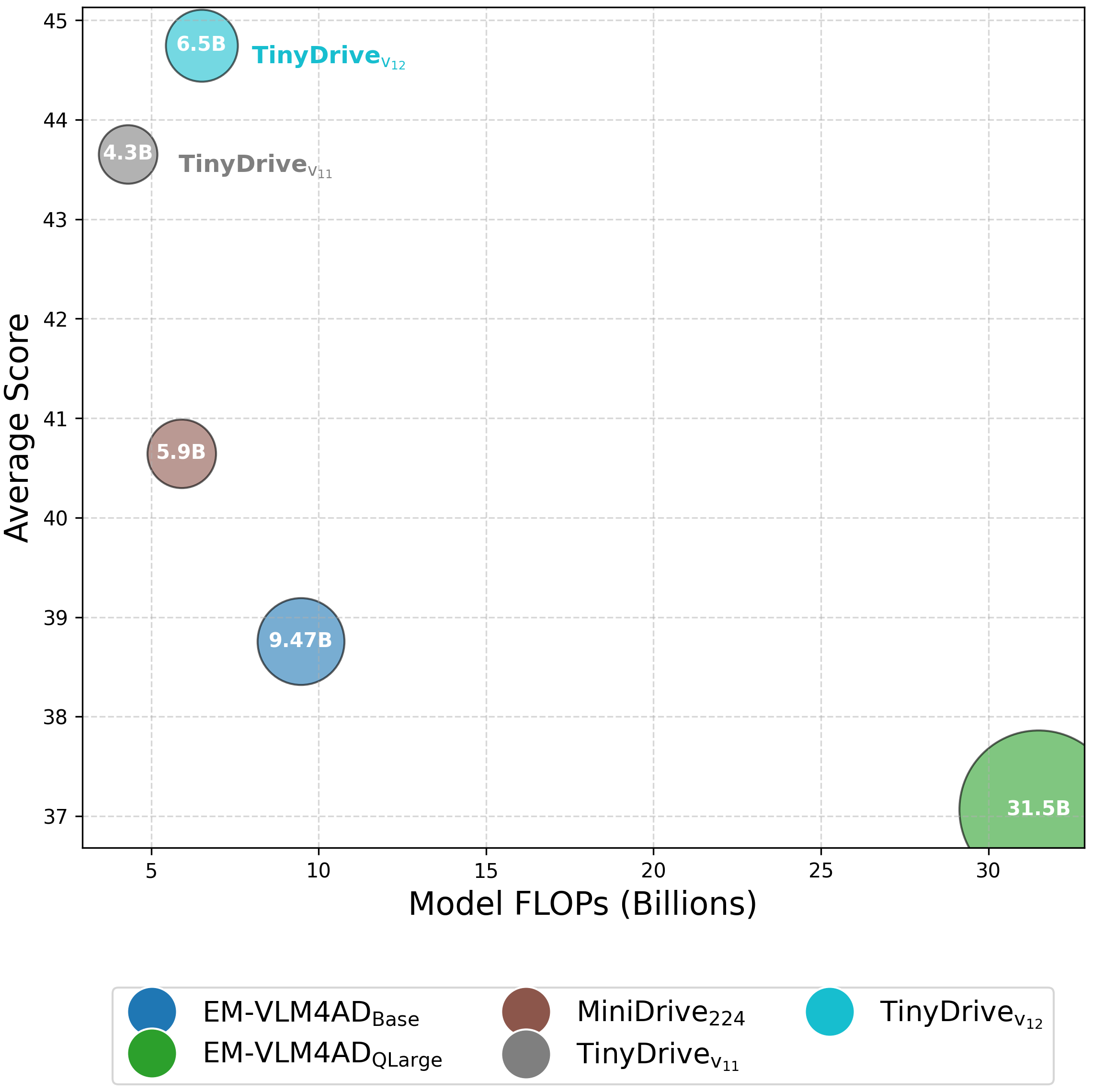}}
    \caption{Comparison of TinyDrive against SOTA models on the DriveLM benchmark, evaluating average language understanding performance against (a) total trainable parameters and (b) total FLOPs. TinyDrive outperforms SOTA models while requiring substantially fewer computational resources.}
    \label{fig_I_I}
\end{figure}

We initially evaluate TinyDrive on our custom-curated VQA dataset. Furthermore, TinyDrive is evaluated under a multi-view VQA task using the DriveLM dataset \cite{sima2024drivelm}. The attained results show that TinyDrive achieves state-of-the-art (SOTA) language understanding performance, while using considerably fewer parameters and FLOPs. Figure \ref{fig_I_I} illustrates the average performance scores of TinyDrive and SOTA models across BLEU-4 \cite{papineni2002bleu}, METEOR \cite{banerjee2005meteor}, ROUGE-L \cite{lin2004rouge}, and CIDEr \cite{vedantam2015cider} with respect to the number of trainable parameters and FLOPs. TinyDrive outperforms SOTA models, while consuming lower computational resources by a significant margin. To this end, the present study contributes to VQA in autonomous driving by:
\begin{itemize}
    \item Constructing a VQA dataset with diverse perception, prediction, planning, and behavior QAs using an experimental self-driving car;
    \item Developing a multiscale vision encoder featuring scale injection, cross-scale gating, and an LGB to capture enriched visual features for VQA and classification;
    \item Proposing a dual-level prioritization for optimized training through a token routing mechanism that only selects informative text tokens to be further processed, and prunes less informative ones;
    \item And constructing a sequence priority buffer to retain and selectively choose sequences that contribute the most into the training.
\end{itemize}

The rest of the paper is organized as follows. Related works are presented in Section \ref{secII}. We discuss the proposed method in detail in Section \ref{secIII}. Experiments and results are illustrated in Section \ref{secIV}, followed by limitations in Section \ref{secV} and conclusions in Section \ref{secVI}.

\section{Related Work}\label{secII}
The development of Transformers has undoubtedly revolutionized natural language processing (NLP) by effectively attending to long-range dependencies in text. Their flexible architecture is suitable for sequence-based natural language tasks which has led to the rise of LLMs. Being trained over massive corpora through encoder (e.g., BERT \cite{devlin2019bert}), decoder (e.g., LLaMA \cite{touvron2023llama}), and encoder-decoder (e.g., T5 \cite{raffel2020exploring}) variants, LLMs capture semantically rich linguistic patterns and can be seamlessly fine-tuned for a wide variety of downstream tasks. Their significant success in language inspired the adaptation of Transformers to vision by the introduction of Vision Transformer (ViT) \cite{dosovitskiy2020image}. By presenting an image as a sequence of patches, ViT can process them similarly to text. This similarity, which can be translated into the convergence of modalities, ultimately led to the emergence of VLMs. By combining ViT and LLMs, VLMs have enabled cross-modal learning to capture image-text relationships for downstream tasks such as image-caption matching through models such as CLIP \cite{radford2021learning} and BLIP-2 \cite{li2023blip}. 

The cross-modal capability of VLMs makes them powerful in VQA tasks, for which understanding both language and vision context is required. This is particularly beneficial in autonomous driving where the ego vehicle can be enabled to answer questions about their immediate environments and interpret complex real-world driving scenes. Benefiting from emergent reasoning enabled through the integrated LLM, a VLM can directly contribute to safer and context-aware decision-making in autonomous vehicles (AVs). In this context, LLaMA-7B \cite{touvron2023llama} has been a foundational language component for several models developed for driving QA tasks in AVs. For instance, Chen \textit{et al.} \cite{chen2024driving} developed a two-phase framework, where they initially project vectorized driving data into embeddings for a frozen LLaMA-7B model. The model is then fine-tuned using LoRA \cite{hu2022lora}. By adopting LLaMA-7B and using CLIP as the vision encoder, DriveGPT4 \cite{xu2024drivegpt4} can process videos of traffic scenes along with prompt text in order to generate answers and also low-level control commands. Building on this, DriveMLM \cite{wang2023drivemlm} integrates multi-view images, LiDAR point clouds, traffic rules, and user commands generated utilizing a simulator. By relying on ViT-g/14 and LLaMA-7B, DriveMLM is able to support complex closed-loop driving. In a different approach, Sha \textit{et al.} \cite{sha2310languagempc} employed a chain-of-thought framework \cite{wei2022chain} for ChatGPT-3.5 for logical explanation and reasoning in driving scenarios. Similarly, Mao \textit{et al.} \cite{mao2023gpt} introduced GPT-Driver, which is built upon GPT-3.5 to address the motioned planning problem through a language modeling by representing inputs and outputs as tokens. More recently, EM-VLM4AD model \cite{gopalkrishnan2024multi} showed improved performance on the DriveLM dataset \cite{sima2024drivelm} compared to previous models relying on billions of parameters. EM-VLM4AD uses a vision encoder, through which vision embeddings of multi-view images were integrated into a single embedding. The constructed unified embedding was then concatenated with text embeddings to generate the input to the language model. Building on this, MiniDrive \cite{zhang2024minidrive} was proposed which includes an effective vision encoder and adapter for multi-view VQA in DriveLM dataset. It incorporates two key components, Feature Engineering Mixture of Experts (FE-MoE) and Dynamic Instruction Adapter (DI-Adapter). The goal of FE-MoE is to project 2D visual features into compact visual token embeddings suitable for the T5 language model, while DI-Adapter dynamically modulates these visual tokens based on instructions extracted from text embeddings. The model achieves SOTA efficiency and superior performance on the DriveLM dataset. 

\section{Methodology}\label{secIII}
In this section, we present the overall architecture of TinyDrive including detailed descriptions of the vision encoder, token routing mechanism, and priority buffer. This will then be followed by a thorough discussion on the fine-tuning strategy.

\begin{figure}
    \centering
    \includegraphics[width=0.7\linewidth]{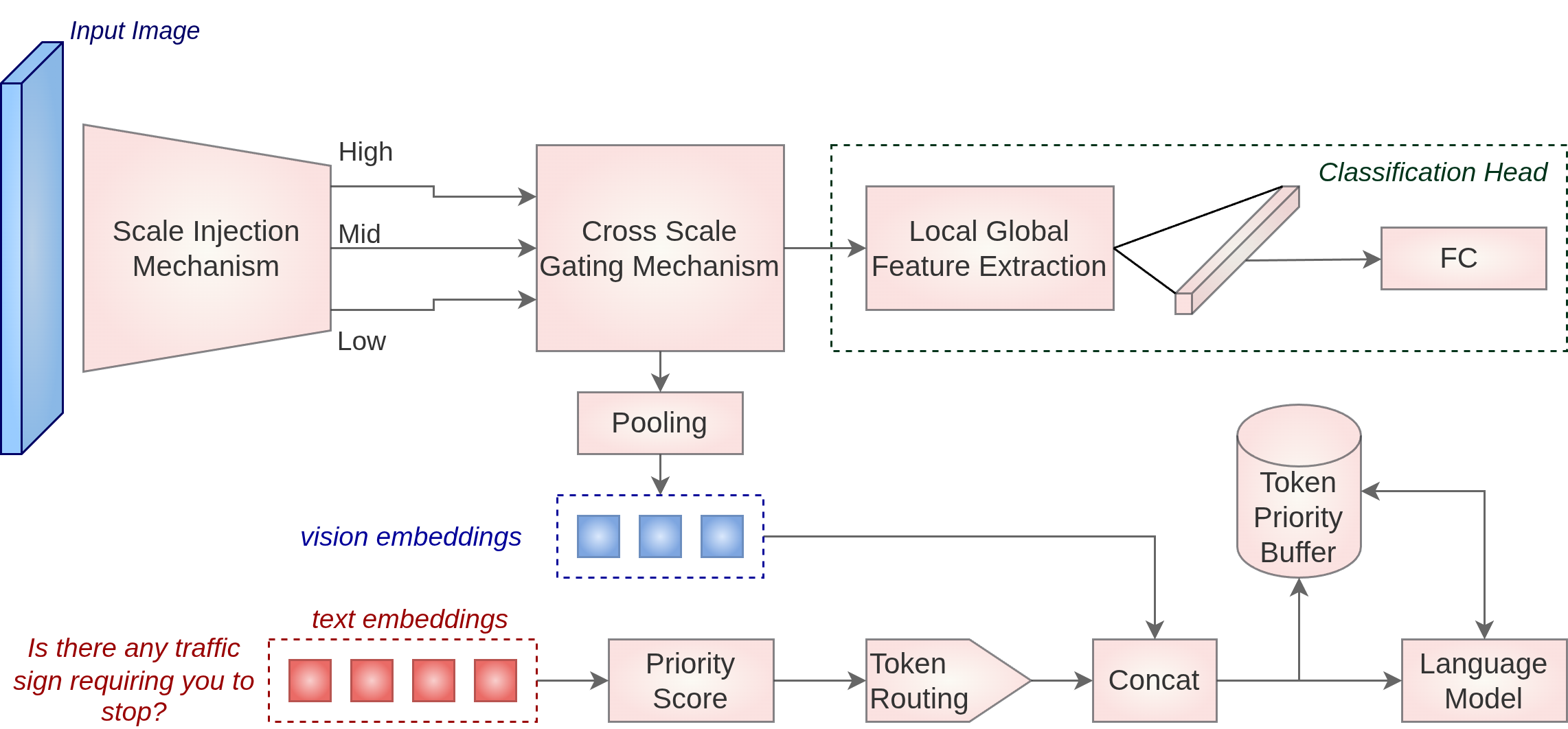}
    \caption{Overall architecture of TinyDrive.}\label{fig_31}
\end{figure}

\subsection{Overall Framework}\label{secIII-I}
Overall framework of TinyDrive is illustrated in Figure \ref{fig_31}. The goal is to weave together a lightweight, multiscale vision encoder with a dynamic pipeline for text token selection in order to generate high-quality VQA on resource-constrained hardware. To achieve this, we first fed an input frame through a three-resolution CNN injection stage (high, mid, low). Outputs of this module are then fused using a cross-scale gating unit before passing through an LGB and a final pooling layer to yield compact vision embeddings. In parallel, we tokenize and score text tokens on the fly using a mixture of gradient magnitudes, uncertainty estimates, and diversity measures in order to flag the most informative $K$ embeddings. We then prune away lower-scoring tokens using the token routing mechanism, where the remaining text vectors are concatenated by the vision features and queued into the sequence priority buffer. During training, sequences with higher aggregate scores are sampled more frequently to fine-tune the T5-based language model.

\subsection{Vision Encoder}\label{secIII-II}
Given $n$ views of a scene, the goal is to learn a unified representation to extract informative vision embeddings. We show the input tensor by $\mathbf{X} \in \mathbb{R}^{b \times n \times 3 \times h \times w}$, in which $b$ is the batch size and each view is an RGB image with spatial dimension $h\times w$. We represent each view by $\mathbf{x}_i \in \mathbb{R}^{b \times 3 \times h \times w}$ for $i \in \{1,2,...,n\}$. The goal is to learn a mapping function $f: \mathbb{R}^{b \times n \times 3 \times h \times w} \rightarrow \mathbb{R}^{b \times d}$ that ultimately projects multi-view input to a compact $d$-dimensional embedding space, where $d = n \times 3 \times c_p$ and $c_p$ shows the projection channels for each scale.

\paragraph{Scale Injection Mechanism (SIM)}
To expand feature dimensionality while preserving spatial resolution, we initially apply a stem network $\mathcal{S}$ to each view $\mathbf{x}_i$ in order to project the input RGB channels into an intermediate representation $\mathbf{F}_i^{\text{stem}}\in \mathbb{R}^{b \times c_s \times h \times w}$ with $c_s$ being the number of stem channels. To capture information at different levels of granularity, we extract features at three spatial scales including high-resolution $\mathbf{F}_i^{\text{high}} \in \mathbb{R}^{b \times c_h \times h \times w}$, mid-resolution $\mathbf{F}_i^{\text{mid}} \in \mathbb{R}^{b \times c_m \times \frac{h}{2} \times \frac{w}{2}}$, and low-resolution $\mathbf{F}_i^{\text{low}} \in \mathbb{R}^{b \times c_l \times \frac{h}{4} \times \frac{w}{4}}$, where the channel dimensions $c_h$, $c_m$, and $c_l$ differ for each branch. Extracted features are then projected to a common channel dimension $c_p$ to ensure channel consistency across scales as $\mathbf{P}_i^{\text{high}} \in \mathbb{R}^{b \times c_p \times h \times w}$, $\mathbf{P}_i^{\text{mid}} \in \mathbb{R}^{b \times c_p \times \frac{h}{2} \times \frac{w}{2}}$, and $\mathbf{P}_i^{\text{low}} \in \mathbb{R}^{b \times c_p \times \frac{h}{4} \times \frac{w}{4}}$.

Multi-scale features provide us with information at different granularities, however, all regions within these scales are not equally informative. To cope with this, we apply channel-wise and spatial attention to each scale's features. Given scale $s \in \{\text{high, med, low}\}$, we apply a squeeze-and-excitation style attention:
\begin{align}
\mathbf{C}_i^s = \mathbf{P}_i^s \odot \sigma\left(\mathcal{F}_{ca}^s\left(\text{GAP}\left(\mathbf{P}_i^s\right)\right)\right),
\end{align}
where $\text{GAP}$ stands for global average pooling, $\mathcal{F}_{ca}^s$ shows channel attention function which is indeed a bottleneck MLP with sigmoid activation $\sigma$, and $\odot$ denotes element-wise multiplication. The spatial attention maps are then extracted  by a different feature combination for each scale as below:
\begin{align}
\mathbf{A}_i^{\text{high}} &= \sigma\left(\mathcal{F}_{sa}^{\text{high}}\left(\left[\text{Avg}(\mathbf{C}_i^{\text{high}}), \text{Max}(\mathbf{C}_i^{\text{high}}), \text{Grad}(\text{Avg}(\mathbf{C}_i^{\text{high}}))\right]\right)\right), \\
\mathbf{A}_i^{\text{mid}} &= \sigma\left(\mathcal{F}_{sa}^{\text{mid}}\left(\left[\text{Avg}(\mathbf{C}_i^{\text{mid}}), \text{Max}(\mathbf{C}_i^{\text{mid}}), \text{LocalVar}(\text{Avg}(\mathbf{C}_i^{\text{mid}}))\right]\right)\right), \\
\mathbf{A}_i^{\text{low}} &= \sigma\left(\mathcal{F}_{sa}^{\text{low}}\left(\left[\text{Avg}(\mathbf{C}_i^{\text{low}}), \text{Max}(\mathbf{C}_i^{\text{low}})\right]\right)\right),
\end{align}
where the channel-wise average and maximum values are computed by $\text{Avg}$ and $\text{Max}$, gradient magnitude by means of Sobel operators are calculated by $\text{Grad}$, $\text{LocalVar}$ is used to show local variance, $\mathcal{F}_{sa}^s$ are convolution operations, and $[\cdot]$ operator represents channel-wise concatenation. The refined features are then as follows:
\begin{align}
\mathbf{\hat{P}}_i^s = \mathbf{C}_i^s \odot \mathbf{A}_i^s, \quad s \in \{\text{high, mid, low}\}.
\end{align}

\paragraph{Cross-Scale Gating Mechanism (CSGM)}
In CSGM, we first upsample the lower resolution features so that they match the spatial dimension of the highest resolution, and, then, concatenate the aligned features and compute adaptive weights using a gating mechanism as in the following:
\begin{align}
\mathbf{F}_i^{\text{concat}} &= \left[\mathbf{\hat{P}}_i^{\text{high}}, \mathbf{\hat{P}}_i^{\text{mid} \uparrow}, \mathbf{\hat{P}}_i^{\text{low} \uparrow}\right] \in \mathbb{R}^{b \times 3c_p \times h \times w}, \\
\mathbf{w}_i &= \text{Softmax}\left(\mathcal{G}(\text{GAP}(\mathbf{F}_i^{\text{concat}}))\right) \in \mathbb{R}^{b \times 3c_p \times 1 \times 1},
\end{align}

where $\mathcal{G}$ is a bottleneck MLP. The gated fusion and fused features can then be computed as below:
\begin{align}\label{eq8}
\mathbf{F}_i^{\text{fused}} = \mathbf{F}_i^{\text{concat}} \odot \mathbf{w}_i \in \mathbb{R}^{b \times 3c_p \times h \times w},\\
\mathbf{e}_i = \text{GAP}(\mathbf{F}_i^{\text{fused}}) \in \mathbb{R}^{b \times 3c_p}.
\end{align}
Following this, our final view-specific vision embeddings are as follows:
\begin{align}
\mathbf{E} = [\mathbf{e}_1, \mathbf{e}_2, \ldots, \mathbf{e}_n] \in \mathbb{R}^{b \times n \times 3c_p}.
\end{align}

\paragraph{Classification Head} 
The generated features by means of CSGM block, $\mathbf{F}_i^{\text{fused}} \in \mathbb{R}^{b \times 3c_p \times h \times w}$ as described in Equation (\ref{eq8}), are further processed by the LGB to generate class logits. Within the LGB, we decompose these features into two paths called \textit{local} and \textit{global}. In the local path, dynamic per-sample local convolutions are employed, while in the global path, we use dilated global convolutions. The concatenated local and global features are downsampled using strided convolutions and max pooling, followed by flattening. The output is then passed through a 2-layer MLP to produce class logits.

\subsection{Token Routing and Priority Buffer}\label{secIII-III}
We propose a dual-level prioritization mechanism to combine token-level and sequence-level selection to optimize VLM training. We define a token routing strategy to retain only informative text tokens within each sample, and employ a sequence priority buffer to sample informative examples.

\paragraph{Token routing mechanism}
At token-level prioritization, we focus on text embeddings and preserve all visual tokens. This is because text tokens have lower information density and higher redundancy compared to visual embeddings. In addition, we aim to reduce computational overhead without degrading VQA performance. In this respect, for each token $t_{i\ell}$ in sequence $i$ at position $\ell \in \{1, \dots, L\}$, we compute a composite importance score:
\begin{align}
    s_{i\ell} = \left[\lambda_m \cdot \left\|\mathbf{e}_{i\ell}\right\|_2 + \lambda_p \cdot \phi(\ell)\right] \cdot \gamma(t_{i\ell}),
\end{align}
where $\mathbf{e}_{i\ell} \in \mathbb{R}^d$ is the embedding magnitude, $\phi(\ell)$ is a position-based weighting function, and $\gamma(t_{i\ell})$ is a padding mask to ensure padding tokens are never selected. Tokens with higher vector magnitudes typically carry stronger semantic signals and tokens in mid-sequence positions are upweighted as below to preserve contextual information:
\begin{equation}
    \phi(\ell) = 
    \begin{cases} 
        1.2, & \text{if } \frac{L}{4} \leq l \leq \frac{3L}{4}, \\
        1.0, & \text{otherwise}.
    \end{cases}
    \end{equation}

The weights $\lambda_m$ and $\lambda_p$ balance each contribution and scores $s_i$ are min-max normalized. We retain only the $K$ most informative text tokens from each sequence.

\paragraph{Sequence-Level Prioritization}
The $i$th sequence $\mathsf{s}_i$ is scored by an informativeness function:
\begin{align}\label{eq13}
    \pi_i = \lambda_\ell \cdot \mathcal{L}_i + \lambda_u \cdot \mathcal{U}_i + \lambda_d \cdot \mathcal{D}_i,
\end{align}
where $\mathcal{L}_i$ is the normalized cross-entropy loss, and $\lambda_\ell$, $\lambda_u$ and $\lambda_d$ control contribution of each metric. We also measure model uncertainty $\mathcal{U}_i$ as inverse of prediction confidence for each sequence as below:
\begin{align}
    \mathcal{U}_i = 1 - \frac{1}{|\mathsf{s}_i|} \sum_{t \in T_i} \max_v P(v|t),
\end{align}
where $|\mathsf{s}_i|$ is the cardinality of the $i$th sequence and $P(v|t)$ is the probability of vocabulary item $v$ for token $t$. Finally, we integrate the diversity metric $\mathcal{D}_i$ to promote sequences dissimilar to others within a training batch as given below:
\begin{align}
   \mathcal{D}_i = 1 - \frac{1}{N-1}\sum_{j \neq i} \textrm{cos}(\mathsf{s}_i, \mathsf{s}_j),
\end{align}
where $\textrm{cos}(\mathsf{s}_i, \mathsf{s}_j)$ is the cosine similarity between embeddings of sequences $i$ and $j$. We preserve a sequence priority buffer, where the uniform minibatch selection is replaced with an importance-weighted sampling. Given the priority scores $\pi_i$ (\ref{eq13}), the probability of the $i$th sequence to be sampled is $\nicefrac{\pi_i^\alpha}{\sum_{j} \pi_j^\alpha}$, where $\alpha \in [0,1]$ controls the prioritization intensity. Since prioritized sampling introduces bias, we correct the bias using importance weights $w_i =( \nicefrac{1}{N} \cdot \nicefrac{1}{P(i)})^{\beta}$, where $N$ is the capacity of the buffer and $\beta$ is the correction parameter. 


\section{Experiments}\label{secIV}
In this section, we first evaluate TinyDrive on our collected dataset called \textit{Rosmaster}. We start with the description of this dataset, followed by presenting and assessing the attained results. We then provide a comparative study with SOTA models on the DriveLM dataset. 

\subsection{Rosmaster Dataset}\label{secIV_I}
To collect this dataset, we used a Yahboom Rosmaster X3 self-driving car in an experimental setup in an indoor lab. The road patterns contains small-scale traffic and road signs (e.g., stop sign, traffic lights). A few sample images captured by the car from this environment are illustrated in Appendix \ref{appI_I}. This dataset contains $2,344$ images, among which we selected $165$ images presenting specific classes that requires the self-driving car to do a maneuver. The reader is referred to Appendix \ref{appI_I} for detailed description of classes. We created a pool of perception, prediction, planning, and behavior questions. Then, for each frame we randomly picked $3$ to $4$ perception, prediction, planning, and behavior QAs from the pool to make the VQA dataset as diverse as possible. In total, we generated $2,279$ QAs for the selected images, as summarized in Table \ref{tab_IV_I}.

\begin{table}
  \caption{Description of datasets used to evaluate TinyDrive. Rosmaster is a custom-curated dataset collected using a Yahboom Rosmaster X3 self-driving car in an experimental setup.}
  \label{tab_IV_I}
  \centering
  \setlength{\tabcolsep}{2pt}
  \begin{tabular}{lccccccc}
    \toprule
    \multirow{2}{*}{Dataset} & \multirow{2}{*}{Type} &  \multirow{2}{*}{\#. Frames} & \multicolumn{4}{c}{QA Pairs} \\
    \cmidrule(lr){4-8}
    & & & Perception & Prediction & Planning & Behavior & Total \\
    \midrule
    Rosmaster & single-view & 165 & 566 & 579 & 563 & 571 & 2,279\\
    DriveLM-nuScenes \cite{sima2024drivelm} & multi-view & 4,871 & 144k & 153k & 146k & - & 443k \\
    \bottomrule
  \end{tabular}
\end{table}

\begin{figure}
    \centering
    \includegraphics[width=0.9\linewidth]{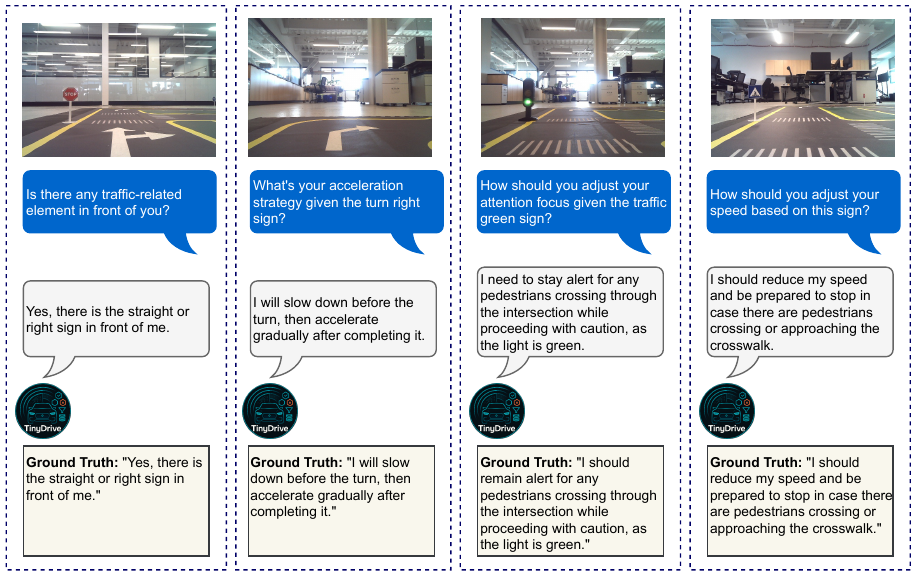}
    \caption{Sample generated answers by TinyDrive$_{v_{12}}$ for VQA on the Rosmaster dataset.}\label{fig_IV_I}
\end{figure}

In this study, we proposed two versions of TinyDrive. The lighter version is TinyDrive$_{v_{11}}$, which uses the proposed vision encoder discussed in Section~\ref{secIII-II} and the T5-tiny language model. The other version is TinyDrive$_{v_{12}}$, which employs the same vision encoder but pairs it with the T5-mini language model. Using these architectures, along with the implementation details discussed in Appendix~\ref{app_II}, the performance of both models on the Rosmaster dataset is summarized in Table~\ref{tab_IV_II}. The results shows that the $v_{12}$ version of TinyDrive consistently outperforms $v_{11}$ across all the language understanding metrics. There has been an improvement of 8.79 points in BLEU-4, 7.94 points in METEOR, 7.15 in ROUGE-L, and 0.88 in CIDEr. The attained high scores by both versions reflects the exceptional performance of TinyDrive for VQA in self-driving cars. A few VQA samples by TinyDrive on the Rosmaster dataset are illustrated in Figure \ref{fig_IV_I}.

\begin{table}
\caption{Performance of TinyDrive$_{v_{11}}$ and TinyDrive$_{v_{12}}$ on the Rosmaster dataset. \textbf{Bold} shows the best outcome.}\label{tab_IV_II}
    \centering
    \begin{tabular}{lcccc}
         \toprule
         Model & BLUE-4$\uparrow$ & METEOR$\uparrow$ & ROUGE-L$\uparrow$ & CIDEr$\uparrow$ \\
         \cmidrule(lr){1-5} 
         TinyDrive$_{v_{11}}$ & 54.37 & 60.70 & 63.59 & 5.201\\
         TinyDrive$_{v_{12}}$ & \textbf{63.16} & \textbf{68.64} & \textbf{70.74} & \textbf{6.081} \\
         \bottomrule
    \end{tabular}
\end{table}

\begin{figure}
\centering
\begin{subfigure}[b]{\textwidth}\centering
   \includegraphics[width=0.95\linewidth]{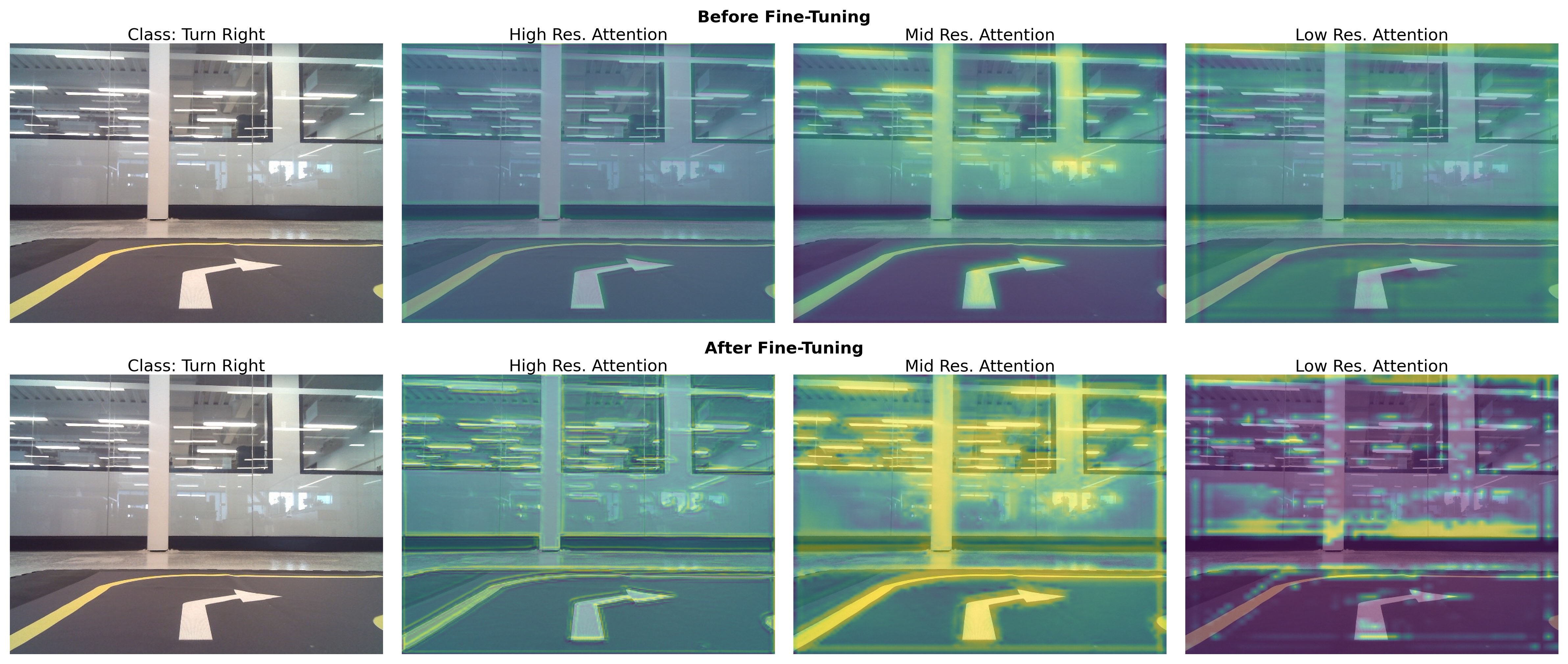}
   \caption{An example of the \textit{turn right} class.}
\end{subfigure}

\begin{subfigure}[b]{\textwidth}\centering
   \includegraphics[width=0.95\linewidth]{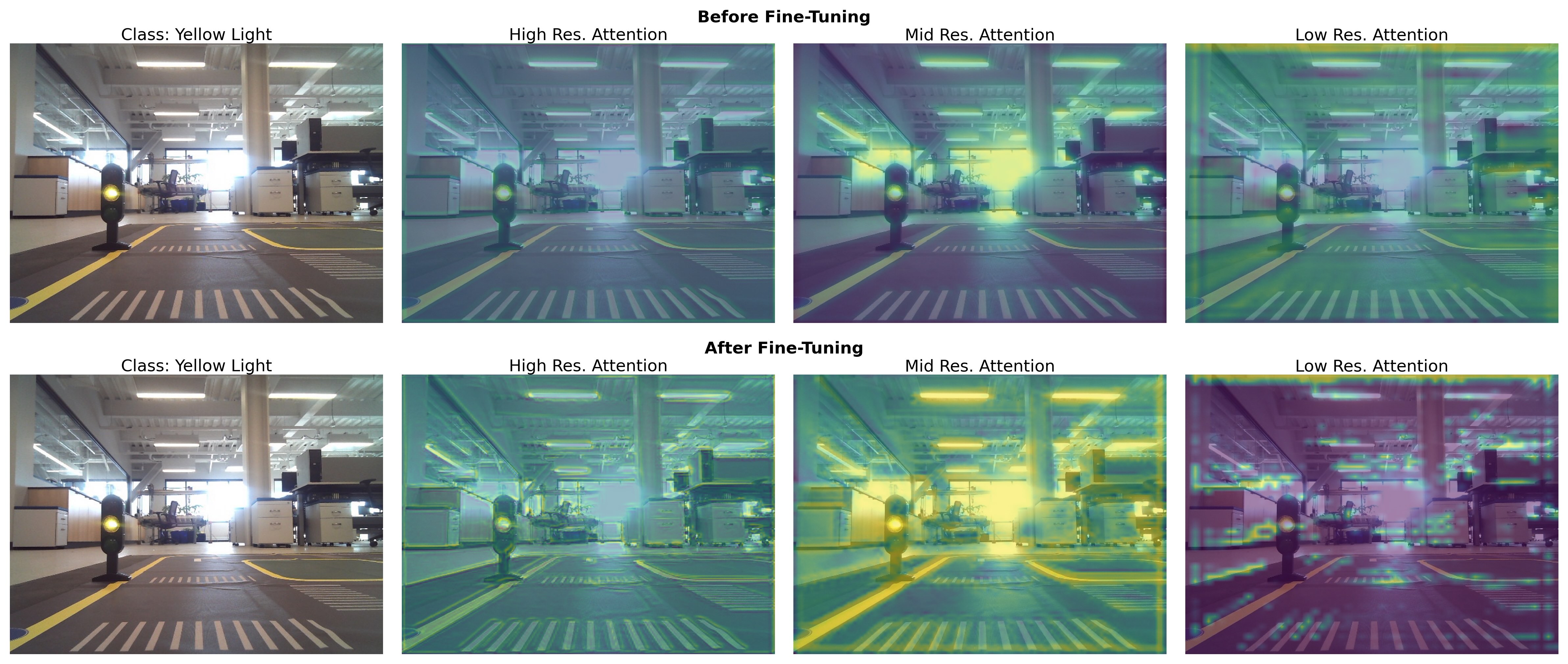}
   \caption{An example of the \textit{yellow light} class.}
\end{subfigure}

\caption{The generated attention maps through the high, mid, and low resolution branches for two classes including (a) \textit{turn right} and (b) \textit{yellow light}. For each sub-figure, the top figure shows the maps before fine-turning the classification head and the bottom illustrates the same but for after fine-tuning.}\label{fig_IV_II}
\end{figure}

During fine-tuning of the language model, we froze the classification head. After fine-tuning, the backbone of the vision encoder was frozen and its classification head was trained. Through this mechanism, attention maps of our vision encoder demonstrated significant improvements after fine-tuning. Figure \ref{fig_IV_II} illustrates the resulting changes in attention weights, where more focus is placed on critical features after training. Furthermore, each resolution branch serves a distinct function. The high-resolution branch captures sign edges, the mid-resolution focuses on sign structure, and the low-resolution provides the scene context. This complementary multi-scale approach enables the vision encoder to achieve 100\% accuracy on the Rosmaster dataset during both training and validation. Out of the total 2,344 images, 10\% were used to validate the vision encoder.

\subsection{DriveLM-nuScenes Dataset}\label{secIV_II}
In this section, we provide a comprehensive comparative study to validate the effectiveness of TinyDrive for multi-view VQA. In this respect, we resort to the DriveLM-nuScenes \cite{sima2024drivelm} dataset, which is a commonly used benchmark to evaluate VLM models for VQA in autonomous driving applications. Details of this dataset are given in Table \ref{tab_IV_I}. There exist 4,871 frames in this dataset, where for each one, a total of 91.4 QA pairs are created on average, resulting in 433k QA pairs in total. The dataset focuses on perception, prediction, and planning tasks, and unlike the Rosmaster dataset, there is no behavior-related QAs. 

\begin{table}
\caption{Architecture of TinyDrive and SOTA models as well as the number of parameters and FLOPs for multi-view VQA on the DriveLM dataset. \textbf{Bold} shows the smallest model and \underline{underline} denotes the second smallest model.}\label{tab_IV_III}
    \centering
    \begin{tabular}{llcc}
         \toprule
         \multicolumn{1}{c}{Model} & \multicolumn{1}{c}{Pretrained Models Used} & Parameters & FLOP count \\
         \cmidrule(lr){1-4} EM-VLM4AD$_{\text{Base}}$ & T5-Base, ViT-b/32 patch embedder & 235M & 9.47B\\
         EM-VLM4AD$_{\text{QLarge}}$ & T5-Large, ViT-b/32 patch embedder & 769M & 31.5B\\
         MiniDrive$_{224}$ & T5-small & 83M & \underline{5.9B}\\
         MiniDrive$_{384}$ & T5-small & 137M & -\\
         TinyDrive$_{v_{11}}$ & T5-tiny & \textbf{16M} & \textbf{4.27B}\\
         TinyDrive$_{v_{12}}$ & T5-mini & \underline{31.8M} & 6.50B\\
         \bottomrule
    \end{tabular}
\end{table}

\begin{table}
\caption{Performance of TinyDrive against SOTA models on the DriveLM-nuScenes dataset. \textbf{Bold} shows the best outcome and \underline{underline} denotes the second best performance.}\label{tab_IV_IV}
    \centering
    \begin{tabular}{lcccc}
         \toprule
         Model & BLUE-4$\uparrow$ & METEOR$\uparrow$ & ROUGE-L$\uparrow$ & CIDEr$\uparrow$ \\
         \cmidrule(lr){1-5}
         EM-VLM4AD$_{\text{Base}}$ & 45.36 & 34.49 & 71.98 & 3.20\\
         EM-VLM4AD$_{\text{QLarge}}$ & 40.11 & 34.34 & 70.72 & 3.10\\
         MiniDrive$_{224}$ & 49.70 & 36.30 & \underline{73.30} & \underline{3.28}\\
         MiniDrive$_{384}$ & 50.20 & 37.40 & \textbf{73.50} & \textbf{3.32}\\
         TinyDrive$_{v_{11}}$ & \underline{54.73} & \underline{49.14} & 67.90 & 2.85\\
         TinyDrive$_{v_{12}}$ &  \textbf{55.82} & \textbf{50.56} & 69.50 & 3.02\\
         \bottomrule
    \end{tabular}
\end{table}

We compare TinyDrive with SOTA models including EM-VLM4AD \cite{gopalkrishnan2024multi} and MiniDrive \cite{zhang2024minidrive}, and their variants. Table \ref{tab_IV_III} compares the architecture of these models in terms of pretrained models used, the number of parameters, and the FLOP count. As it can be observed from this table, TinyDrive$_{v_{11}}$ has the lowest parameters and FLOP count, with 16M and 4.27B, respectively. These numbers increase to 31.8M and 6.50B for the TinyDrive$_{v_{12}}$, which is built upon the same vision encoder but leverages the T5-mini language model. The lower number of parameters is mainly due to the lightweight design of our vision encoder matched with our token routing mechanism.

The quantitative comparison of TinyDrive against SOTA models is represented in Table \ref{tab_IV_IV}. Results demonstrate remarkable trade-off between efficiency and performance for both TinyDrive versions compared to SOTA models on the DriveLM-nuScenes dataset. Regardless of having only 16M parameters, TinyDrive$_{v_{11}}$ outperform any variants of MiniDrive and EM-VLM4AD in terms of BLEU-4 and METEOR. In the same vein, TinyDrive$_{v_{12}}$ with only 31.8M parameters, a 76.8\% reduction versus MiniDrive$_{384}$, can achieve superior performance on key metrics. Particularly, TinyDrive$_{v_{12}}$ also outperforms all competitors on BLEU-4 and METEOR with scores of 55.82 and 50.56, respectively, which represents an 11.1\% improvement in BLEU-4 and a significant 35.4\% improvement in METEOR compared to the previous best model (MiniDrive$_{384}$). When averaging across all four metrics presented in Table \ref{tab_IV_IV}, TinyDrive$_{v_{12}}$ achieves a mean score of 44.73 and outperforms MiniDrive$_{384}$ (41.11), MiniDrive$_{224}$ (40.70), EM-VLM4AD$_{\text{QLarge}}$ (37.07), and EM-VLM4AD$_{\text{Base}}$ (38.75), as it was initially presented in Figure \ref{fig_I_I} in the Introduction. Even the smaller TinyDrive$_{v_{11}}$, with 16M parameters, shows superior performance across all metrics while requiring only 4.27B FLOPs, demonstrating a reduction of 99\% compared to EM-VLM4AD$_{\text{Base}}$ (9.47B FLOPs). Therefore, results of this comparative study highlight the significant efficiency-performance gains of TinyDrive on driving scene understanding tasks. This is of particular importance when it comes to deploying VLMs for VQA in resource-constrained self-driving vehicles.  

\section{Limitations}\label{secV}
The first generation of TinyDrive, i.e., $v_{11}$ and $v_{12}$, considers text tokens only within the token-level prioritization. In this respect, the constructed score concerns text embeddings only, and there is no measure of how the selected tokens are related to vision embeddings. Even though at the sequence-level prioritization we take both text and vision into account, however, a score function that relates text and vision at the first level of prioritization could potentially contribute more to training efficiency. This could be done by cross-modal relevance scoring mechanisms such as cross-attention-based scoring, similarity-based filtering, or a fusion module that learns a joint representation for each text token conditioned on the vision input.

\section{Conclusion}\label{secVI}
This paper introduced TinyDrive, an efficient VLM for VQA in autonomous driving. The overarching goal of TinyDrive was enhancing the efficiency-performance trade-off. This was achieved by proposing a lightweight, multiscale vision encoder using CNNs. To further improve training, a novel dual-level prioritization was proposed, through which the top text tokens are initially selected, and then, the most informative sequences are more frequently selected for fine-tuning the language model. Results on a custom-generated Rosmaster dataset, and the public DriveLM benchmark demonstrate that our model achieves significant performance improvement over the SOTA models, while utilizing fewer trainable parameters. 



\appendix

\section{Rosmaster Dataset}\label{appI_I}
\begin{figure}
    \centering
    \includegraphics[width=\linewidth]{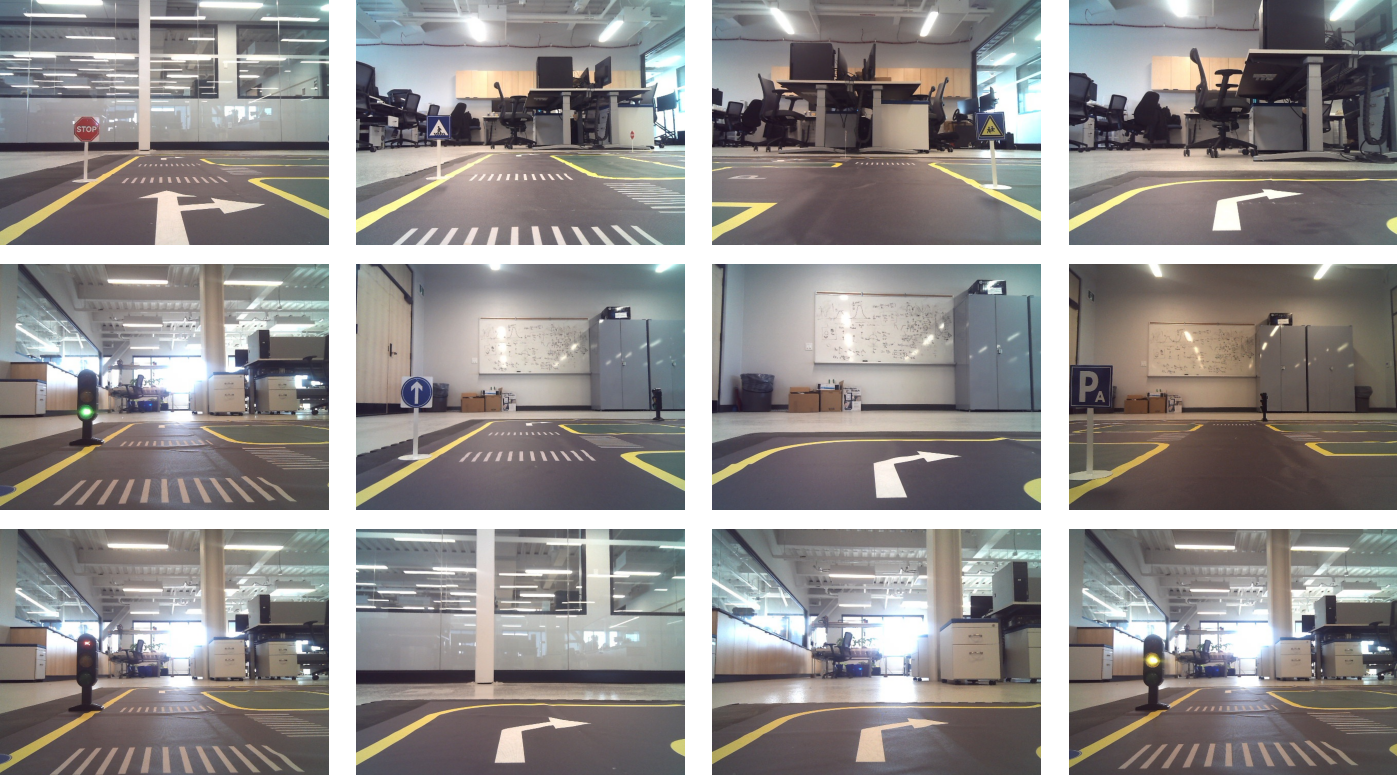}
    \caption{Sample images from the Rosmaster dataset. Images are captured using an RGB camera mounted on the self-driving car navigating through a driving map within a laboratory.}\label{fig_appI}
\end{figure}

Rosmaster dataset was collected using a Yahboom Rosmaster X3 self-driving car on a road map within our laboratory. The map is accompanied with small-scale traffic signs such as stop sign, one-way road, traffic lights, as well as marked signs on the road to denote traffic directions such as turn right and pedestrian crossing markings. The car is equipped with a depth camera and a LiDAR sensor, all running on a Jetson NANO 4GB board. It supports both ROS1 and ROS2, and is programmable using Python. We used the RGB-depth camera mounted on the car, and have programmed the car to navigate through the map and collect RGB images of resolution 640$\times$480 from all over the map. Examples of such images are illustrated in Figure \ref{fig_appI}.

To send control commands to the car, we then labeled the images using the LabelImg tool \cite{tzutalin2015labelimg}, which is an open-source software under the MIT License. The labeled images were used to train the classification head of the vision encoder so that it follows the specific maneuvers facing each traffic and road sign. For example, the car has to come to a full stop when facing a stop sign, stay there for five seconds, and then move forward. In this respect, the Rosmaster dataset contains 11 different classes including \{\textit{straight or right}, \textit{stop sign}, \textit{turn right}, \textit{one-way road}, \textit{green light}, \textit{yellow light}, \textit{red light}, \textit{pedestrian crossing}, \textit{school zone}, \textit{parking}\}. An additional class called \textit{no object} is also included, where the vehicle maintains its speed and trajectory without executing any particular maneuvers. 

For fine-tuning the language model, the created VQA dataset is divided into training, validation, and test sets with proportions of 0.7, 0.2, and 0.1, respectively. The same proportions are applied for fine-tuning the classification head of the vision encoder.

\begin{table}
  \caption{Hyperparameters used in the TinyDrive implementation.}\label{tab_app_II}
  \centering
  \setlength{\tabcolsep}{4pt}
  \begin{tabular}{llccc}
    \toprule
    \multirow{2}{*}{Component} & \multirow{2}{*}{Parameter} & \multirow{2}{*}{Symbol} & \multicolumn{2}{c}{Value} \\
    \cmidrule(lr){4-5}
    & & & DriveLM & Rosmater \\
    \midrule
    \multirow{5}{*}{Token Generation} & Image Resolution & $h\times w$ & $224\times224$ & $224\times224$\\
     & Question Max Length & - & $75$ & $75$\\
     & Answer Max Length & - & $128$ & $128$\\
     & Text Embedding Dimension & $L$ & $256$/$384$ & $256$/$384$\\
     & Vision Embeddings Dimension & $n\times c_p$ & $144$ & $24$\\
     \cmidrule(lr){1-5}
     \multirow{9}{*}{Vision Encoder} & Stem Channels & $c_s$ & $8$ & $8$\\
     & High-Resolution Channels & $c_h$ & $16$ & $16$\\
     & Mid-Resolution Channels & $c_m$ &  $32$ & $32$\\
     & Low-Resolution Channels & $c_l$ &  $64$ & $64$\\
     & Common Projection Channels & $c_p$ &  $8$ & $8$\\
     & Number of Classes & - & N/A & 11\\
     & FC Units & - & $64$ & $64$\\
     & FC Dropout & - & $0.2$ & $0.2$\\
     & Number of views & $n$ & $6$ & $1$\\
     \cmidrule(lr){1-5}
     \multirow{9}{*}{Fine-Tuning (language)} & Epochs & - & $15$ & $15$\\
     & Batch Size & $b$ & $4$ & $4$\\
     & Learning Rate & - & $10^{-3}$ & $10^{-3}$\\
     & Weight Decay & - & $10^{-2}$ & $10^{-2}$\\
     & Top K Text Tokens & $K$ & $64$ & $64$\\
     & Prioritization Intensity & $\alpha$ & $0.6$ & $0.6$\\
     & Correction Parameter & $\beta$ & $0.4$ & $0.4$\\
     & $\beta$ annealing rate & - & $0.001$ & $0.001$\\
     & Buffer Capacity & - & $50,000$ & $5,000$\\
     & Loss Weight & $\lambda_{\ell}$ & $0.5$ & $0.5$\\
     & Uncertainty Weight & $\lambda_u$ &  $0.3$ & $0.3$\\
     & Diversity Weight & $\lambda_d$ & $0.2$ & $0.2$\\
     & Magnitude Weight & $\lambda_m$ & $1.0$ & $1.0$\\
     & Positional Weight & $\lambda_p$ & $1.0$ & $1.0$\\
     \cmidrule(lr){1-5}
     \multirow{2}{*}{Fine-Tuning (vision)} & Epochs & - & N/A & $100$\\
     & Batch Size & $b$ & N/A & $16$\\
    \bottomrule
  \end{tabular}
\end{table}

\section{Implementation Details}\label{app_II}
This section provides a comprehensive overview of the implementation details for TinyDrive. Summary of hyperparameters used in our experiments is given in Table \ref{tab_app_II}. It is worth noting that all the reported hyperparameters were optimized through trial and error, where we selected the best-performing configuration. Furthermore, it should be noted that all experiments were conducted on a PC running Ubuntu 24.04, equipped with an AMD Ryzen 9 7950X CPU (16 cores, 32 threads), 64GB of RAM, a single NVIDIA RTX 4090 GPU with 24GB memory, and a 2 TB SSD.

\subsection{Token Generation}
To prepare the input data by combining visual and textual characteristics, we initially process images, where each one is resized to a resolution of $224\times224$ pixels. The image tensor of shape $[1, 3, 224, 224]$ is then processed using the vision encoder detailed in Section \ref{secIII-II} to generate vision embeddings of size $[1,144]$ (number of views $\times$ number of branches $\times$ common channel dimension $=6\times 3\times 8=144$). For the text data, we use T5 tokenizer from T5-tiny/mini model to tokenize each question with a maximum length of 75 tokens by considering the padding and truncations, yielding a text embedding of shape $[1,256]$ ($[1,384]$ for T5-mini). Vision and text embeddings are then concatenated to form a $144+256=400$ ($144+384=528$ for T5-mini) dimension input.

\subsection{Vision Encoder}
Architecture of the vision encoder is comprehensively discussed in Section \ref{secIII-II}. The goal is to process images at multiple scales to capture both fine-grained and contextual features, through the following step-by-step implementation.

\paragraph{Stem} The input image of shape $[B,3,224,224]$ is first processed using a stem that consists of a 2D convolution with a $3\times 3$ kernel, stride 1, and padding 1, followed by batch normalization and ReLU activation. The number of stem channels is 8.

\paragraph{Multi-Scale Branches} In the high-resolution branch, the input is processed using a convolution layer (kernel size $3\times 3$, stride 1, padding 1) and number of channels is 16. In the mid-resolution branch, a $2\times 2$ max-pooling with stride 2 is followed by a convolution layer to produce 32 channels. In the low-resolution branch, we initially apply two $2\times 2$ max-pooling operations with stride 2, which are then followed by a convolution layer and produce 64 channels.

\paragraph{Projection} The output of each branch is projected into a common channel dimension of 8 by means of $1\times 1$ convolutions, which result in a total of $24$ channels after concatenation. 

\paragraph{Scale Injection} We initially apply channel and spatial attention to the high-, mid-, and low-resolution feature maps. Within the channel attention, we use adaptive average pooling and convolutional layers to reduce channels by a factor of 4, which is followed by sigmoid activation. For spatial attention, we use different kernel sizes (3 for high, 5 for mid, and 7 for low), and also integrate gradient magnitude for high-resolution and local variance for medium resolution features.

\paragraph{Cross-Scale Gating} Multi-scale features are fused by interpolating mid- and low-resolution maps to the high-resolution size. We then concatenate feature maps of each branch, and apply a gating mechanism to reduce the dimension of 24 to 16 before passing them through the softmax. 

\paragraph{Local Global Block} We combine local dynamic convolutions with kernels per samples in the local path, and dilated convolutions in the global path, and perform downsampling to reduce dimension to 12. 

\paragraph{Pooling and Fully-Connected Layer} Fused features are passed through two convolutional layers with $3\times 3$ kernels and stride 2 to reduce the channels to 6. This is then followed by $4\times 4$ max-pooling with stride 4, and adaptive average pooling to reduce the spatial dimension to $1\times 1$. The flattened features of size $6\times 7\times 7=294$ are then fed into a two-layer MLP with 64 units and an $0.2$ dropout rate, where the output layer then generates the class logits of size 11.

\subsection{Fine-Tuning Process}
We employ a dropout rate of 0.2 for both general and attention dropout in the language model. To prepare the data, we wrapped the train, validation, and test tokens using the custom \texttt{PrioritizedVQADataset} instances, for which the buffer capacity is set to 50,000, and $\alpha=0.6$ and $\beta=0.4$. We pad input IDs, attention masks, and labels to the maximum length in each batch using a custom collation function. We use AdamW optimizer with a learning rate of $10^{-3}$ for the language model, and a smaller learning rate of $10^{-4}$ for the vision encoder. The weight decay is set to $10^{-2}$. We employ an exponential learning rate scheduler with $\gamma=0.9$, a batch size of 4, and training epochs of 15. In all experiments, we keep 64 text embeddings through the proposed token-routing mechanism. 

\subsection{Dual-Level Prioritization}
The buffer capacity is set to 50,000 (5,000 for the Rosmaster dataset), the prioritization exponent is $\alpha=0.6$, and importance sampling exponent is $\beta=0.4$, which is annealed toward 1 with a rate of $0.001$. The weights in token scores are $\lambda_m=\lambda_p=1.0$, and for the sequence-level prioritization, the weights are set to $\lambda_{\ell}=0.5$, $\lambda_u=0.3$, and $\lambda_d=0.2$. During fine-tuning, the informativeness scores are calculated for each batch and are updated every $5$ batches.

\end{document}